\documentclass[letterpaper, 10 pt, conference]{ieeeconf}  
\usepackage{amsmath}
\usepackage{graphicx}
\usepackage{cite}
\usepackage{hyperref}
\hypersetup{hidelinks,pdfauthor={},pdfsubject={},pdfkeywords={}}
\hypersetup{hidelinks}
\usepackage{amssymb}

\IEEEoverridecommandlockouts                              

\title{\LARGE \bf
To Coordinate or Not to Coordinate? Learning to Swim with Eight Soft Arms
}

\author{
Junkai Chen$^{1,\dagger}$, Dehan Yuan$^{1,3,\dagger}$, Yiyuan Zhang$^{1,2,*}$, Yu Duan$^{1}$,\\
Fan Shi$^{2}$, Dixia Fan$^{3}$, and Cecilia Laschi$^{1}$, \textit{Fellow, IEEE}%
\thanks{$^{\dagger}$Junkai Chen and Dehan Yuan contributed equally to this work. $^{*}$Corresponding author: Yiyuan Zhang (e-mail: yiyuan.zhang@u.nus.edu).}%
\thanks{$^{1}$Department of Mechanical Engineering, College of Design and Engineering, and Advanced Robotics Centre, National University of Singapore, Singapore.}%
\thanks{$^{2}$Department of Electrical and Computer Engineering, National University of Singapore, Singapore.}%
\thanks{$^{3}$Department of Electronic and Information Engineering, School of Engineering, Westlake University, Hangzhou, China.}%
}

\begin{document}

\maketitle
\thispagestyle{empty}
\pagestyle{empty}

\begin{abstract}
Does effective swimming with multiple underactuated soft arms require explicit coordination between arms? We hypothesize that an eight-arm swimmer can control its swimming speed and direction by independently learning each arm's actuation amplitude under a shared rhythmic drive, without prescribed inter-arm phase coupling. To test this hypothesis, we compare three reinforcement learning-based control strategies using proximal policy optimization (PPO): (1) learning arm-specific actuation amplitudes without inter-arm phase coupling, (2) learning actuation amplitudes with prescribed inter-arm phase coupling, and (3) directly learning actuator commands through end-to-end PPO without an explicit rhythmic structure. Across 200 episodes with randomized three-dimensional velocity commands, the controller without phase coupling satisfies the tracking criterion in 169 episodes, compared with 117 for the phase-coupled controller and 28 for end-to-end PPO. Straight-line swimming and commanded direction-change trials further demonstrate more accurate velocity tracking without phase coupling. These findings suggest that effective eight-arm swimming does not require prescribed inter-arm phase coupling, but can instead be achieved through learned arm-specific amplitude modulation within a rhythmic control structure.
\end{abstract}

\section{INTRODUCTION}
\label{sec:introduction}

Underwater robots use multiple soft appendages to produce propulsion and interact with their surroundings \cite{c2,c3}. Compliant propulsion has been demonstrated through cyclic arm strokes in octopus-like swimmers \cite{c13,c1} and through rotating soft flagellar modules \cite{c30}. Designs that combine swimming with grasping further show how the same appendages can support locomotion and environmental interaction \cite{c31,c33}. The resulting motion depends jointly on actuator inputs, arm elasticity, and hydrodynamic loading \cite{c34,c10}. This interaction makes it difficult to translate a commanded swimming velocity into individual arm inputs.

Existing controllers address this problem through mechanical models, prescribed movement patterns, and feedback. Model-based control uses mechanical descriptions to relate actuation to motion \cite{c25}, while predefined arm trajectories have enabled multi-arm swimming \cite{c13}. A study of a four-arm webbed swimmer examined how web geometry and arm actuation patterns influence swimming performance \cite{c32}. For underwater crawling, motion primitives combined with closed-loop planning have enabled a five-legged soft robot to move toward a goal \cite{c35}. These studies motivate asking how much of the movement pattern should be prescribed when the objective is swimming-velocity tracking.

Central pattern generators (CPGs) provide rhythmic drive and a means of prescribing phase relationships between actuator outputs \cite{c4}. Reinforcement learning (RL) has been used to optimize a coupled-CPG controller for undulating-fin propulsion \cite{c22} and to learn swimming patterns in robotic fish \cite{c23}. Supplying a rhythm and prescribing inter-arm phase coupling are nevertheless distinct design choices. The question for the present swimmer is whether prescribed inter-arm phase coupling is useful for velocity tracking, or whether learned arm-specific amplitude modulation under a shared rhythm without that coupling is more effective.

We hypothesize that an eight-arm swimmer can control its swimming speed and direction through learned arm-specific amplitude modulation under a shared rhythmic drive, without prescribed coupling. To test this hypothesis, we introduce a feedback control architecture called Non-coordinated PPO, which uses proximal policy optimization (PPO) to learn each arm's actuator-command amplitude gain and one common rhythm frequency. We compare it with Coordinated PPO, which imposes a circumferential phase pattern through a coupled CPG, and Direct PPO, which is an intuitive approach for solving the problem.  Building on passively morphing soft-arm propulsion \cite{c1} and modeling approach \cite{c10}, we evaluate the controllers on an eight-arm swimmer robot using randomized three-dimensional velocity commands and representative straight-line and direction-change maneuvers.

\section{Methodology}
\label{sec:method}

\subsection{Robot Model and Simulation}
\label{subsec:robot}

\begin{figure*}
    \centering
    \includegraphics[width=\linewidth]{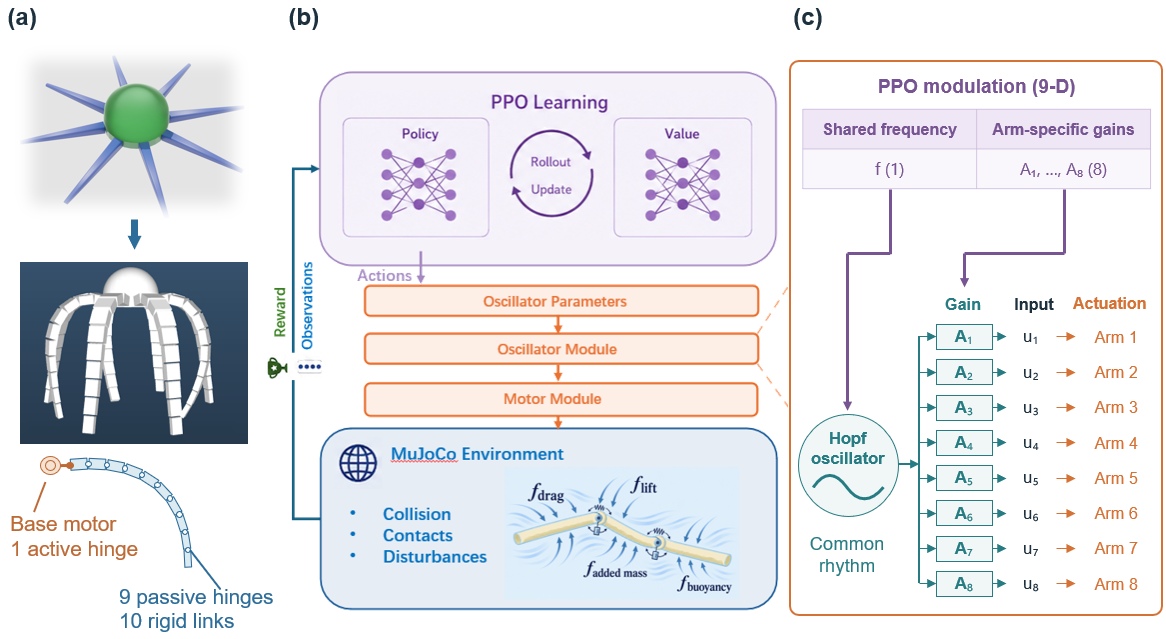}
    \caption{Non-coordinated PPO for an underactuated eight-arm underwater robot. (a) Idealized morphology, multibody realization, and a bent-arm schematic. The orange motor symbol marks the actuated base hinge; ten rigid links and nine passive hinges approximate arm compliance. (b) PPO learning and closed-loop control. Observations from MuJoCo drive the policy, while rewards guide updates to the policy and value networks. The policy modulates a shared Hopf oscillator and the actuator mapping. (c) Nine-dimensional policy modulation consists of one common frequency $f$ and eight arm-specific amplitude gains $A_1,\ldots,A_8$. The frequency drives a single oscillator, whose common waveform is distributed to all arm channels. Each gain scales its channel before a fixed control midpoint and clipping map it to a bounded actuator input $u_i$. The hydrodynamics inset in (b) is a conceptual illustration generated using OpenAI's image-generation tool.}
    \label{fig:general}
\end{figure*}

We study a simulated underactuated underwater robot with a rigid central body and eight radially arranged soft arms. Fig.~\ref{fig:general}(a) shows the idealized morphology (top), its multibody realization (middle), and an articulated-arm schematic (bottom); the arrow denotes the approximation of the deformable arms by articulated segments. The central body comprises a cylindrical plate of radius 30~mm and thickness 12~mm with an upper dome. Eight identical arms attach at the plate perimeter, with nominal azimuths $\psi_i=\pi/8+i\pi/4$, $i=0,\ldots,7$. Each arm extends approximately 160~mm from base to tip in the straight reference configuration and tapers toward its tip. Table~\ref{tab:robot_model} summarizes the geometric and joint parameters.

Each arm is represented by ten serial rigid segments, connected to the central body through one actively driven base hinge and nine passive hinges along the arm. All hinge axes of an arm are parallel to its local tangential axis, so bending is confined to the plane containing that arm's radial direction and the central body's axis. This plane moves with the central body. The central body is free to translate and rotate in three dimensions. The model therefore has 80 internal rotational degrees of freedom, comprising eight actuated and 72 passive joints, together with six degrees of freedom for the floating body.

Soft-arm compliance is represented by elastic bending distributed across the passive hinges; the individual segments remain rigid. Each passive hinge has a zero rest angle and a linear spring--damper response to base actuation and modeled fluid loading. The model specifies MuJoCo's \texttt{springdamper} pair as a time constant of 5.0~s and a damping ratio of 0.02. MuJoCo computes the corresponding stiffness and damping for each joint from its inertia in the reference configuration. Segment masses and inertias are inferred from the geometry and the specified density of $1250~\mathrm{kg/m^3}$.

\begin{table}[t]
\centering
\caption{Geometry and joint parameters of the simulated underactuated underwater robot}
\label{tab:robot_model}
\setlength{\tabcolsep}{4pt}
\renewcommand{\arraystretch}{1.1}
\begin{tabular}{ll}
\hline
Parameter & Value \\
\hline
Central plate radius / thickness & 30 / 12~mm \\
Arm count / angular spacing & 8 / $45^\circ$ \\
Arm base-to-tip reach & Approximately 160~mm \\
Segment count / segment length & 10 per arm / 15~mm \\
Proximal segment width / thickness & 20 / 10~mm \\
Distal segment width / thickness & 10 / 5~mm \\
Geometry density & $1250~\mathrm{kg/m^3}$ \\
Base-joint angular limits & $[-0.523,1.57]$~rad \\
Passive-joint angular limits & $[-0.1,1.57]$~rad \\
Base-joint damping & $0.2~\mathrm{N\,m\,s/rad}$ \\
Joint armature & $10^{-4}~\mathrm{kg\,m^2}$ \\
Passive spring time constant / ratio & $5.0$~s / $0.02$ \\
\hline
\end{tabular}
\end{table}

The robot is simulated in MuJoCo with a physics time step of
$\Delta t_{\mathrm{sim}}=0.002~\mathrm{s}$. Ten physics steps are executed per policy action, resulting in a control frequency of $50$~Hz. Gravitational acceleration is set to zero to approximate neutral buoyancy, while hydrodynamic resistance is approximated using MuJoCo's inertia-based fluid-force model. The fluid viscosity parameter is left at its default value. The fluid density is set to $1000~\mathrm{kg/m^3}$. Each arm-base joint is driven by a direct torque actuator with a constant gain and a force limit,
\begin{equation}
    \tau_i=\operatorname{clip}_{[-10,10]}(3u_i),
\end{equation}
where $\tau_i$ is the applied joint torque (N\,m) and the control input is bounded by
$u_i\in[-0.523,1.046]$. The control-input bounds differ from the mechanical base-joint angular limits in Table~\ref{tab:robot_model}: $u_i$ sets actuator torque, not a desired joint angle. The fluid model applies resistance based on each body's equivalent inertia box; it does not resolve the surrounding flow or wake interactions between arms.

\subsection{Non-coordinated PPO}
\label{subsec:decoupled}

Non-coordinated PPO tests swimming without a prescribed phase-coupled CPG pattern. A single Hopf oscillator supplies a shared rhythmic drive to all eight arms, while their actuator-command amplitude gains are independently adjustable. Its state
\begin{equation}
\begin{aligned}
\dot{x}&=\alpha(\mu-x^2-y^2)x-\omega y,\\
\dot{y}&=\alpha(\mu-x^2-y^2)y+\omega x,
\end{aligned}
\qquad \omega=2\pi f.
\label{eq:hopf}
\end{equation}
We set $\alpha=20$ and $\mu=1$, giving the continuous-time stable
limit-cycle radius $r_0=\sqrt{\mu}=1$. The oscillator is reset to
$\mathbf z(0)=[1,0]^\mathsf T$ and integrated at each physics step
using the explicit forward Euler rule,
$\mathbf z^{n+1}=\mathbf z^n+\Delta t_{\mathrm{sim}}\dot{\mathbf z}^{n}$.

Fig.~\ref{fig:general}(b) and Fig.~\ref{fig:general}(c) summarizes Non-coordinated PPO. A centralized policy maps the velocity command and measured robot state to eight actuator-command amplitude gains and one common frequency. The shared periodic signal is scaled independently for each actuator channel before application to the arm-base torque actuators.

At each 50-Hz policy update, PPO outputs eight normalized amplitude
parameters and one frequency parameter,
\begin{equation}
\begin{split}
\mathbf a_t & = [a^A_{1,t},\ldots,a^A_{8,t},a^f_t]\in[-1,1]^9, \\
A_{i,t} & = 0.6(a^A_{i,t}+1), \quad f_t=1.2+0.6a_t^f.
\end{split}
\label{eq:rhythm_actions}
\end{equation}
Thus $A_{i,t}\in[0,1.2]$ and $f_t\in[0.6,1.8]$~Hz.
With $u_m=(u_{\min}+u_{\max})/2$ and
$u_h=(u_{\max}-u_{\min})/2$, the implemented oscillator-to-actuator
mapping is
\begin{equation}
u_{i,t}=\operatorname{clip}_{[u_{\min},u_{\max}]}
\left(u_m+u_h A_{i,t}
\operatorname{clip}_{[-1,1]}\!\left(\frac{x_t}{r_0}\right)\right).
\label{eq:decoupled_output}
\end{equation}
Amplitude and frequency commands remain fixed over each 20-ms policy
interval, while the oscillator state and actuator command are updated every 2~ms.
All eight actuators therefore receive the same rhythmic phase while the policy can vary each arm's contribution independently at every control interval. The shared reference fixes relative arm timing while leaving the magnitude of each arm command adaptable to the target velocity. Non-coordinated PPO uses one centralized feedback policy with eight independently adjustable output gains, rather than eight independently trained policies.

\subsection{Coordinated PPO}

\label{subsec:coupled}
Coordinated PPO introduces prescribed inter-arm phase coordination through a coupled-oscillator CPG. It uses one Hopf oscillator
$\mathbf z_i=[x_i,y_i]^\mathsf T$ per arm, connected in a
nearest-neighbor ring. The prescribed phase increment $\phi=2\pi/8$
corresponds to the $45^\circ$ angular spacing of the radially arranged
arms. This is an intentionally simple geometry-matched phase-wave baseline. 
The phase-biased coupling is
\begin{equation}
\dot{\mathbf z}_i=\mathbf F(\mathbf z_i,f)
+\frac{K}{d_i}\sum_{j\in\mathcal N_i}
 w_{ij}\left[\mathbf R((i-j)\phi)\mathbf z_j-\mathbf z_i\right],
\label{eq:coupled_osc}
\end{equation}
where $\mathbf F$ denotes the intrinsic dynamics in
Eq.~\eqref{eq:hopf}, $\mathcal N_i$ denotes the two adjacent
oscillators, $d_i=2$, $K=2~\mathrm{s^{-1}}$, and $\phi=\pi/4$.
The planar rotation matrix is
\begin{equation}
\mathbf R(\theta)=
\begin{bmatrix}\cos\theta&-\sin\theta\\
\sin\theta&\cos\theta\end{bmatrix}.
\end{equation}
Each oscillator uses the same intrinsic Hopf parameters and the same
forward Euler integrator. The prescribed initial phases form a
one-cycle phase progression around the eight arms:
$\mathbf z_i(0)=r_0[\cos(i\phi),\sin(i\phi)]^\mathsf T$.
The policy still controls only eight amplitude gains and one common
frequency through Eq.~\eqref{eq:rhythm_actions}. The output for arm $i$
uses the corresponding oscillator component $x_i$ in place of $x$ in
Eq.~\eqref{eq:decoupled_output}.

Both oscillator implementations update at the 2-ms physics rate, whereas
policy actions are updated at 50~Hz. Note that the coupled oscillators
are initialized on the prescribed phase-locked solution and receive an
identical frequency modulation, without independent phase perturbations
or oscillator feedback from mechanical sensors. On the phase-locked
manifold $\mathbf z_i(t)=\mathbf R(i\phi)\mathbf z(t)$, every coupling
difference $\mathbf R((i-j)\phi)\mathbf z_j-\mathbf z_i$ is zero.
In the present experiments, the coupling term serves as a
potential phase-restoring mechanism, the baseline is therefore effectively a fixed circumferential phase pattern with learned per-arm amplitudes and one learned common frequency. It cannot adapt individual phase offsets or arm-specific frequencies within the common nine-dimensional action budget. The experiment therefore compares two timing constraints, not the independent effect of nonzero coupling strength.

\subsection{Direct PPO}
\label{subsec:e2e}
Direct PPO provides an end-to-end comparison without an explicit rhythmic drive. It bypasses the oscillator and
directly outputs eight normalized actuator actions,
$\mathbf a_t\in[-1,1]^8$, mapped by
\begin{equation}
u_{i,t}=u_m+u_h a_{i,t}.
\label{eq:e2e_control}
\end{equation}
These commands are held constant for ten physics steps (20~ms).
In contrast, the oscillator controllers generate periodically varying
actuator commands during each policy interval. This baseline assesses the overall value of the proposed rhythm-guided actuation pipeline relative to direct policy-level actuator commands. The controllers are trained using the same reward formulation and coefficients, while their actuator-command generation differs: the oscillator policies update rhythmic outputs at the physics rate and Direct PPO holds each action for one policy interval.

\subsection{Velocity-Tracking Task}
\label{subsec:system}

The commanded quantity is the world-frame linear velocity of the
robot's entire kinematic-subtree center of mass, as returned by the MuJoCo \texttt{subtreelinvel} sensor:
$\mathbf{v}^{*}=[v_x^{*},v_y^{*},v_z^{*}]^{\mathsf T}$.
At the start of each training episode, a command is independently sampled
from

\begin{equation}
v_x^{*},v_y^{*}\in[-0.06,0.06],\qquad
v_z^{*}\in[0,0.075]~\mathrm{m/s},
\label{eq:velocity_domain}
\end{equation}

and held constant for the episode.
A curriculum linearly increases the
half-width of each sampling interval about its center from approximately
30\% to 100\% over the first 500 episodes \emph{per parallel environment}. Each training episode lasts 10~s (500 policy steps).

The robot-state observation comprises the velocity command and error,
measured world-frame center-of-mass velocity, body-frame angular velocity,
world vertical direction expressed in the body frame, orientation quaternion,
and latest actuator command:

\begin{equation}
\begin{gathered}
\mathbf o^{\mathrm{robot}}_t = \left[
\mathbf v^*_t, \,
\mathbf v^*_t-\mathbf v_t, \,
\mathbf v_t, \,
\boldsymbol\omega^b_t, \,
\mathbf R(\mathbf q_t)^\mathsf T\hat{\mathbf z}^w, \,
\mathbf q_t, \,
\mathbf u_t^{\mathrm{cache}}
\right] \\
\in\mathbb R^{27}.
\end{gathered}
\label{eq:robot_obs}
\end{equation}

The controllers additionally receive a cached policy-action history,
while oscillator-based controllers also observe their oscillator states. In the current implementation, the action cache is updated after constructing
the next observation, so the observed action history has an additional
one-step delay relative to the most recently applied policy action. The resulting observation dimensions are 38 for Non-coordinated PPO and 52 for Coordinated PPO. Direct PPO is padded with three constant-zero channels to form a 38-dimensional observation. Input observations are normalized using running statistics and clipped to $[-10,10]$.

\subsection{Reward and PPO Training}
\label{subsec:training}

Let $\mathbf e_t=\mathbf v^*-\mathbf v_t$ and
$e_t=\|\mathbf e_t\|_2$, where $\mathbf v_t$ is the unfiltered
measured velocity. The velocity-tracking term uses
$\sigma=0.035~\mathrm{m/s}$, and errors above
$\epsilon_v=0.012~\mathrm{m/s}$ incur an additional linear penalty.
The implemented per-step reward is
\begin{equation}
\begin{aligned}
r_t = {} & \exp\!\left(-\frac{e_t^2}{2\sigma^2}\right) - 5\max(e_t-\epsilon_v,0) + 2(e_{t-1}-e_t) \\
& + 0.1\,\operatorname{clip}(\hat{\mathbf z}^{\mathsf T}\mathbf R_b^w(t)\hat{\mathbf z},0,1) \\
& - 0.02\,\frac{\|\mathbf a_t-\mathbf a_{t-1}\|_2^2}{d_a} \\
& + \frac{0.5}{1+\exp\left[-(\epsilon_v-e_t^{\mathrm{EMA}})/0.01\right]} \\
& - 0.5\|\mathbf v_t-\mathbf v_{t-1}\|_2 - w_F F_t,
\end{aligned}
\label{eq:reward_full}
\end{equation}
where $d_a=9$ for the oscillator controllers and $d_a=8$ for the
end-to-end controller. The uprightness term is the cosine of the angle
between the body and world vertical axes, clipped to $[0,1]$.
The EMA velocity is updated as
\begin{equation}
\begin{aligned}
\widetilde{\mathbf v}_t & = \beta\mathbf v_t+(1-\beta)\widetilde{\mathbf v}_{t-1}, \\
\beta & = 1-\exp(-\Delta t_{\mathrm{ctrl}}/0.5), \quad e_t^{\mathrm{EMA}}=\|\mathbf v^*-\widetilde{\mathbf v}_t\|_2.
\end{aligned}
\label{eq:ema_velocity}
\end{equation}
The rolling velocity-fluctuation measure $F_t$ is the root-mean-square
three-dimensional deviation from the mean velocity in the latest
1-s window ($N=50$), set to zero until the window is full:

\begin{equation}
\begin{gathered}
F_t = \sqrt{\frac{1}{N}\sum_{k=t-N+1}^{t} \|\mathbf v_k-\overline{\mathbf v}_{t,N}\|_2^2}, \\
\overline{\mathbf v}_{t,N} = \frac{1}{N}\sum_{k=t-N+1}^{t}\mathbf v_k.
\end{gathered}
\label{eq:f_t_definition}
\end{equation}

The term $\|\mathbf v_t-\mathbf v_{t-1}\|_2$ is a
step-to-step velocity-increment penalty; it is not a time-normalized
physical jerk. All three controllers use the same reward terms and coefficient settings.

An episode is marked successful if the Euclidean error of the \emph{mean} velocity over the final two seconds is
at most $0.012~\mathrm{m/s}$:
\begin{equation}
\left\|\mathbf v^*-
\frac{1}{100}\sum_{k=T-99}^{T}\mathbf v_k\right\|_2
\leq0.012~\mathrm{m/s}.
\label{eq:success}
\end{equation}
The criterion is evaluated at the end of a completed episode, rather
than requiring the instantaneous velocity to remain within the
threshold at every step due to the intermittent motion characteristics of the robot configuration.

All policies are optimized using PPO, using separate actor and critic
multilayer perceptrons with two 256-unit ReLU hidden layers each. The
training configuration specifies $5\times10^7$ total control-level
environment interactions, a rollout length of 2048 steps per environment,
minibatches of 4096 samples, and six optimization epochs per rollout.
The discount factor is $\gamma=0.995$, the GAE parameter is
$\lambda=0.95$, the PPO clipping ratio is 0.2, the value-function
coefficient is 0.5, and the maximum gradient norm is 0.5. The learning
rate decays linearly from $3\times10^{-4}$ to $10^{-5}$, while the
entropy coefficient decays linearly from 0.005 to zero. Training
configuration uses 64 parallel environments.

Observations and rewards are normalized through \texttt{Vec\-Normalize},
with clipping thresholds of 10; evaluation uses frozen observation
statistics and disables reward normalization.


\section{Results and Discussion}
\subsection{Randomized Velocity Tracking}
\label{subsec:random_velocity}

\begin{figure}[t]
    \centering
    \includegraphics[width=\linewidth]{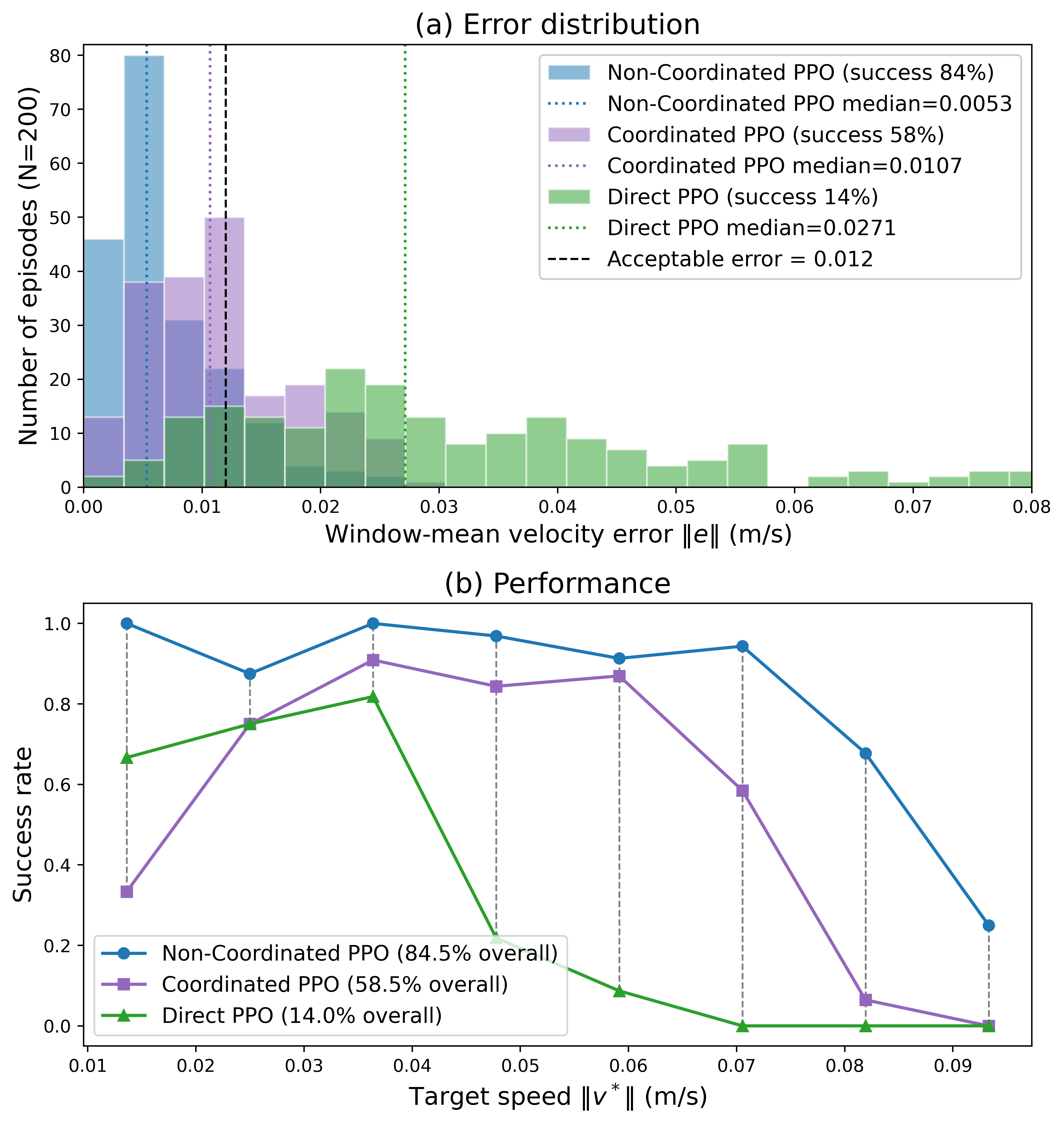}
    \caption{Velocity-tracking performance over 200 randomly sampled commands from the training target-velocity domain. (a) Distribution of the error between the commanded velocity and the measured mean velocity over the final 2\,s of each trial. Colored dotted lines mark controller-specific medians; the black dashed line indicates the success threshold of $0.012\,\mathrm{m/s}$. (b) Tracking success rate grouped by commanded speed $\|\mathbf v^*\|_2$. Overall success rates are shown in the legend.}
    \label{fig:fig_9d_vs_coupled_vs_e2e}
\end{figure}

We first assess prescribed inter-arm coordination by evaluating the three controllers on 200 target velocities sampled
from the same target domain used in the final stage of training. For a
constant target $\mathbf v^*$, the final-window mean velocity and its
tracking error are defined as

\begin{equation}
\begin{gathered}
\overline{\mathbf v}_{\mathrm{final}} =\frac{1}{N}\sum_{k=T-N+1}^{T}\mathbf v_k, \\
E_{\mathrm{final}}=\left\|\overline{\mathbf v}_{\mathrm{final}}-\mathbf v^*\right\|_2,
\end{gathered}
\label{eq:results_final_error}
\end{equation}

where $N=100$ corresponds to the last 2\,s at the 50-Hz control rate.
A trial is classified as successful when
$E_{\mathrm{final}}\leq0.012\,\mathrm{m/s}$. This is a
\emph{mean-velocity} criterion.

As shown in Fig.~\ref{fig:fig_9d_vs_coupled_vs_e2e}(a), Non-coordinated PPO
achieves a median final-window error of $0.0053\,\mathrm{m/s}$ and a
success rate of $84.5\%$ (169/200 trials). Coordinated PPO obtains a median
error of $0.0107\,\mathrm{m/s}$ and a success rate of $58.5\%$ (117/200),
whereas Direct PPO achieves a success rate of $14.0\%$ (28/200 trials).

Thus, Non-coordinated PPO outperforms Coordinated PPO by 26 percentage points in this in-distribution test. The comparison with Direct PPO also reveals a large performance difference under the implemented training configurations.
Figure~\ref{fig:fig_9d_vs_coupled_vs_e2e}(b) breaks down success rates by commanded speed. The reported results indicate that the Non-coordinated PPO controller maintains relatively high success across low and moderate command speeds, while success decreases toward the upper portion of the sampled range.
Coordinated PPO shows stronger degradation at higher speeds, notably around and above $0.07$--$0.08\,\mathrm{m/s}$, whereas Direct PPO loses tracking success earlier as commanded speed increases.

Because the randomized commands span different speeds and directions, these results support effective commanded velocity tracking without the prescribed inter-arm phase pattern across the sampled domain. The lower success of Coordinated PPO shows that the tested phase progression did not improve tracking. Both rhythmic methods outperform Direct PPO under the evaluated conditions, supporting the use of a rhythmic drive.

\subsection{Straight-Line Swimming}
\label{subsec:straight}

\begin{figure}[t]
    \centering
    \includegraphics[width=\linewidth]{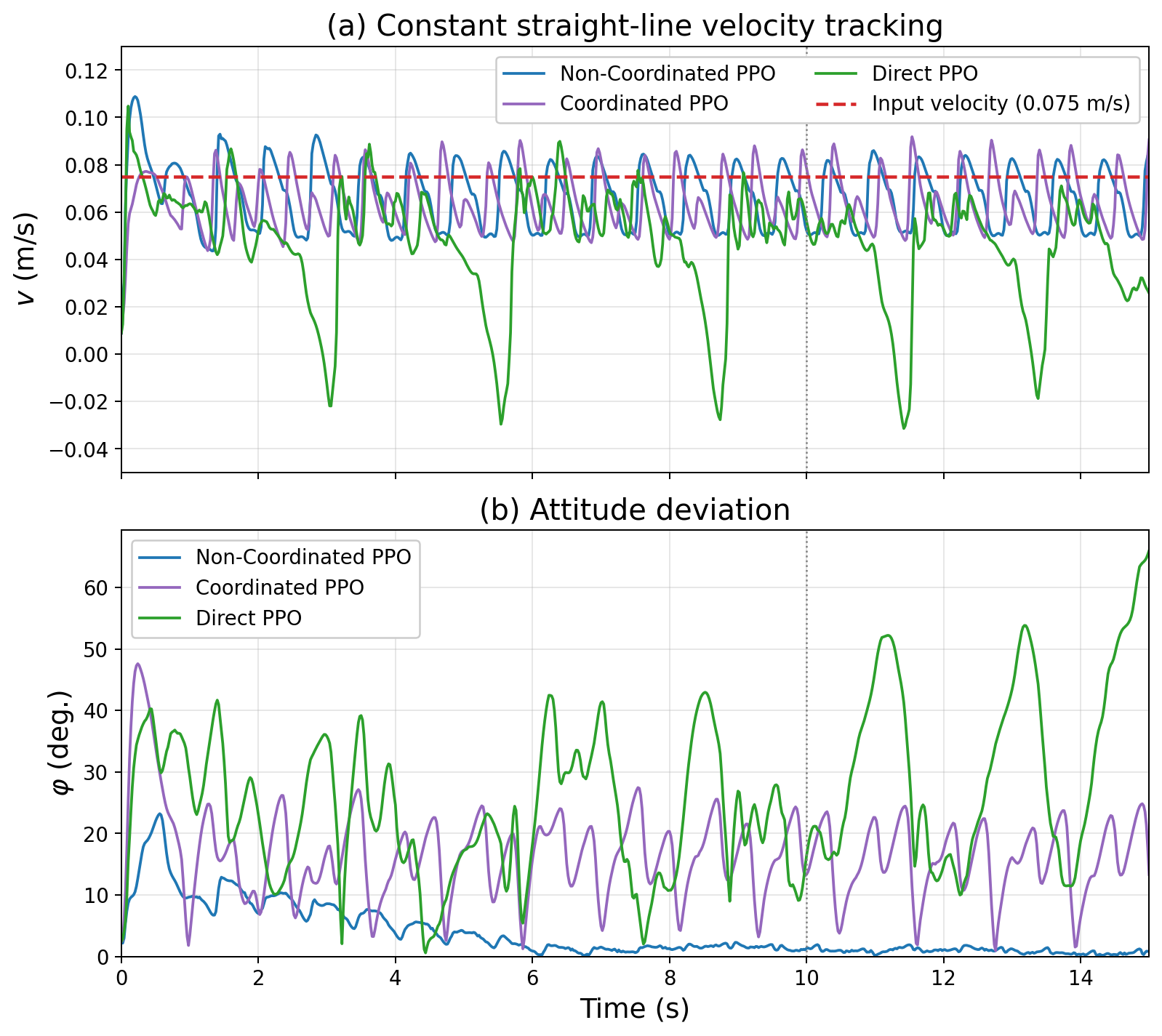}
    \caption{Responses to a constant world-frame velocity command~$\mathbf v^*=[0,0,0.075]^\mathsf T~\mathrm{m/s}$ over a 15-s trial. 
    (a)~Signed velocity component projected onto the commanded direction; the dashed red line denotes $0.075\,\mathrm{m/s}$. 
    (b)~Body tilt~$\varphi$, defined as the angle between the body and world $+z$ axes. This angle measures tilt relative to the world vertical and does not capture rotation about that axis.}
    \label{fig:constant_velocity_tracking}
\end{figure}

\begin{figure}[t]
    \centering
    \includegraphics[width=0.8\linewidth]{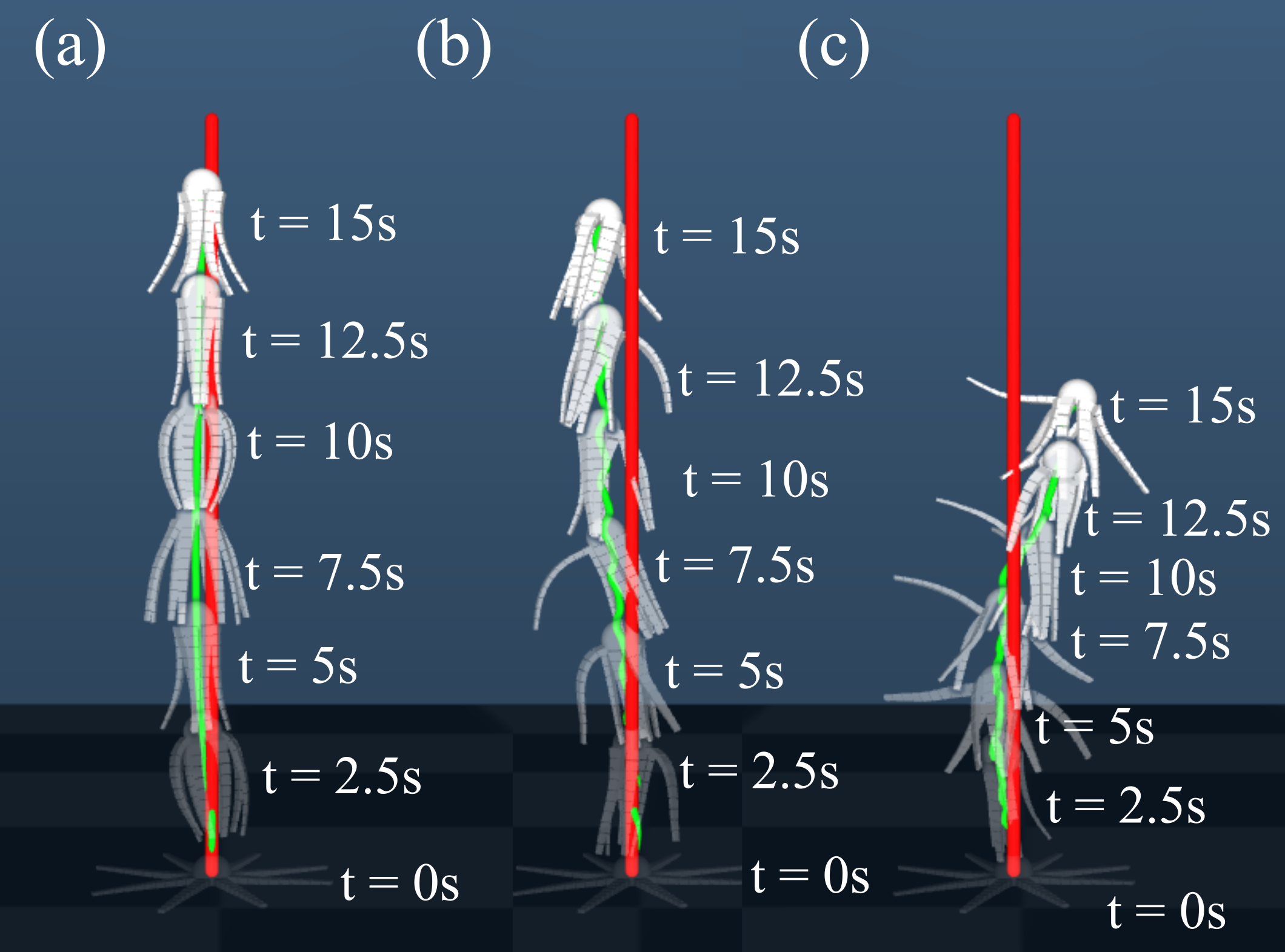}
    \caption{Straight-line swimming comparison for (a) Non-coordinated PPO,
    (b) Coordinated PPO, and (c) Direct PPO. Robot configurations are overlaid
    at 2.5-s intervals from $t=0$ to $15\,$s. The red line is the reference
    trajectory obtained by integrating the constant velocity command, and
    the green line is the measured trajectory of the central platform origin.}
    \label{fig:straight}
\end{figure}

To examine transient and steady motion under a fixed command, the policies
are evaluated for 15\,s with
$\mathbf v^*=[0,0,0.075]^\mathsf T\,\mathrm{m/s}$. Figure~\ref{fig:constant_velocity_tracking}(a)
plots the velocity component along the commanded direction,
$v_{\parallel}(t)=\mathbf v(t)^\mathsf T\mathbf v^*/\|\mathbf v^*\|_2$.
This projection characterizes motion along the commanded direction.
Figure~\ref{fig:constant_velocity_tracking}(b)
reports the body-tilt angle,

\begin{equation}
\begin{gathered}
\varphi(t) = \frac{180^\circ}{\pi}\arccos\!\left(
\operatorname{clip}\!\left(\hat{\mathbf z}^{\mathsf T}
\mathbf R(t)\hat{\mathbf z},-1,1\right)\right), \\
\hat{\mathbf z} = [0,0,1]^\mathsf T.
\end{gathered}
\label{eq:body_tilt_results}
\end{equation}

where $\mathbf R(t)$ maps body-frame vectors into the world frame.

In the illustrated trial, both oscillator-based controllers exhibit oscillatory velocity responses about their forward motion, while Direct PPO shows larger interruptions to forward progression, including intervals of negative projected velocity. The body-tilt traces further indicate a smaller sustained deviation for Non-coordinated PPO than for the other controllers, whereas Coordinated PPO retains more pronounced tilt oscillations and Direct PPO displays larger irregular deviations. Figure~\ref{fig:straight} complements the velocity projection with spatial
trajectories of the central platform origin. The reference path is obtained
by integrating the commanded velocity from that origin's initial position and the policy tracks whole-robot center-of-mass velocity. Therefore the two
points need not move identically when the compliant arms deform. Non-coordinated PPO follows the
straight reference path more closely in the displayed rollout. Coordinated PPO shows greater lateral deviation, whereas Direct PPO exhibits substantial path departure and body reorientation.
 The advantage of Non-coordinated PPO is visible in path adherence and body tilt, rather than in projected forward speed alone.

\subsection{Right-Angle Velocity-Direction Transition}
\label{subsec:turning}

\begin{figure}[htbp]
    \centering
    \includegraphics[width=0.6\linewidth]{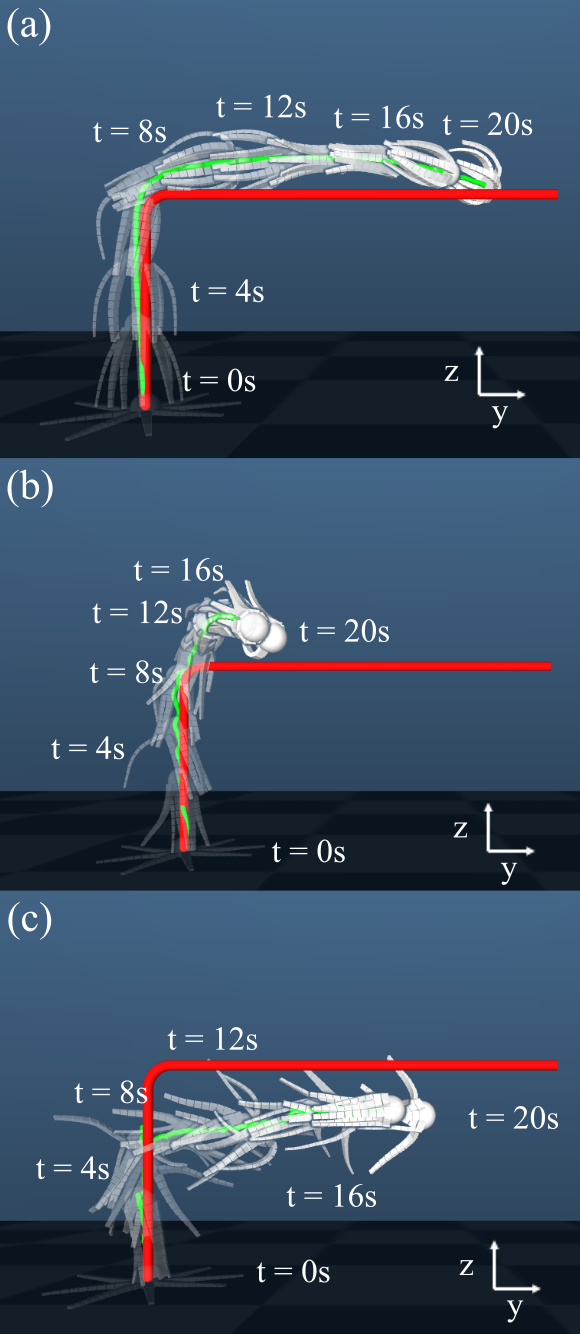}
    \caption{Commanded 90$^\circ$ velocity-direction transition in the world
    $y$--$z$ plane. The command points along $+z$ for the first 5\,s, rotates toward $+y$ over the next 1\,s, and then remains along $+y$. (a)--(c) show Non-coordinated PPO, Coordinated PPO, and Direct PPO. Red and green curves denote the reference path and central-platform trajectory, respectively; robot snapshots illustrate changes in configuration. The command does not prescribe a 90$^\circ$ body-yaw rotation.}
    \label{fig:turnright}
\end{figure}

\begin{figure}[htbp]
    \centering
    \includegraphics[width=0.9\linewidth]{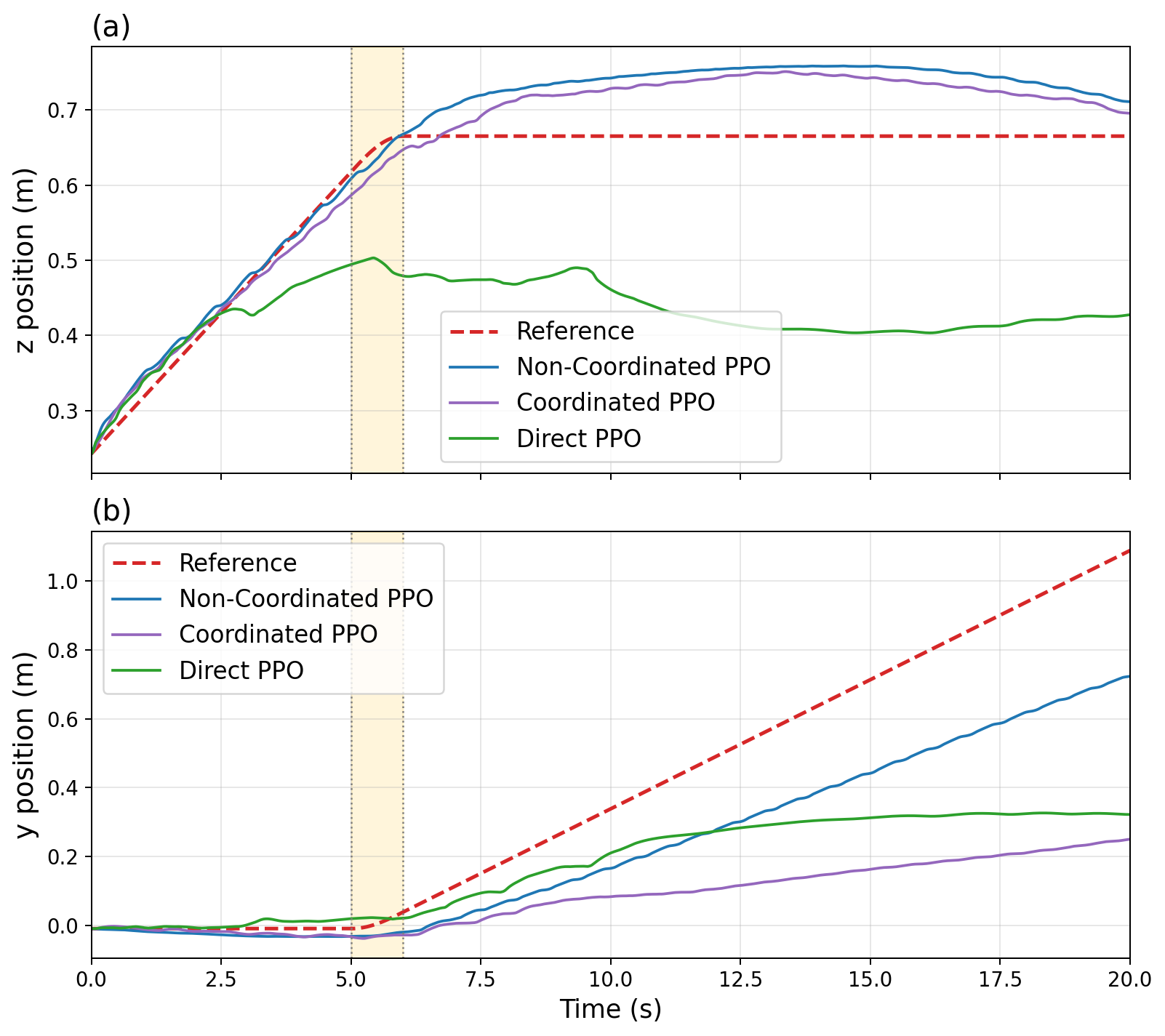}
    \caption{Trajectory components during the commanded velocity-direction
    transition: (a) world-frame $z$ position and (b) world-frame $y$
    position. Red dashed curves indicate the integrated reference velocity
    command, and shading identifies the 5--6-s command-transition interval.}
    \label{fig:right_angle_turning_tracking}
\end{figure}

The third experiment examines adaptation to a time-varying velocity command.
For the first 5\,s, the robot is commanded to move along world $+z$ at
$0.075\,\mathrm{m/s}$. During $5\leq t<6\,$s, the commanded direction
rotates uniformly by 90$^\circ$ in the world $y$--$z$ plane while its
magnitude remains constant. Thereafter, the reference points along world
$+y$ for the remainder of the 20-s trial. In the transition interval,

\begin{equation}
\begin{gathered}
\mathbf v^*(t)=0.075\, [0,\sin\theta(t),\cos\theta(t)]^\mathsf T, \\
\theta(t)=\frac{\pi}{2}(t-5).
\end{gathered}
\label{eq:turning_command_results}
\end{equation}

Figures~\ref{fig:turnright} and~\ref{fig:right_angle_turning_tracking}
compare the reference path with the resulting central-platform motion. In the displayed rollout, both oscillator controllers follow the initial segment more closely than Direct PPO. Following the command transition, Non-coordinated PPO exhibits a stronger response in the newly commanded $+y$
direction and remains closer to the reference overall.
Coordinated PPO shows less post-transition lateral progression, while Direct PPO moves farther in $+y$ than Coordinated PPO but
exhibits larger deviations from the full reference path. Thus, the
Non-coordinated PPO controller provides the most favorable qualitative trade-off between directional response and trajectory adherence in this maneuver. This directional response further supports the hypothesis: the robot changes its swimming direction without the prescribed inter-arm phase wave while retaining a shared rhythm.

The post-turn command $v_y^*=0.075\,\mathrm{m/s}$ lies beyond the
lateral training interval $[-0.06,0.06]\,\mathrm{m/s}$; hence, this
maneuver partly evaluates responses outside the training-command range. The result shows that Non-coordinated PPO can still perform readily well when running an unseen task.

\subsection{Control Implications}
\label{subsec:discussion_scope}

The result indicates that the prescribed inter-arm phase coordination was not required for the demonstrated swimming velocity control. Non-coordinated PPO achieved the highest tracking success through arm-specific modulation under a common rhythm. The lower success of Coordinated PPO indicates that the tested phase progression did not improve control, although both methods retained rhythmic actuation and independently adjustable arm gains. Both methods outperformed Direct PPO under the implemented training setup, indicating that providing a rhythmic movement structure can be more effective than learning actuator inputs directly for this underactuated swimming task.

\begin{figure}[!ht]
    \centering
    \includegraphics[width=0.8\linewidth]{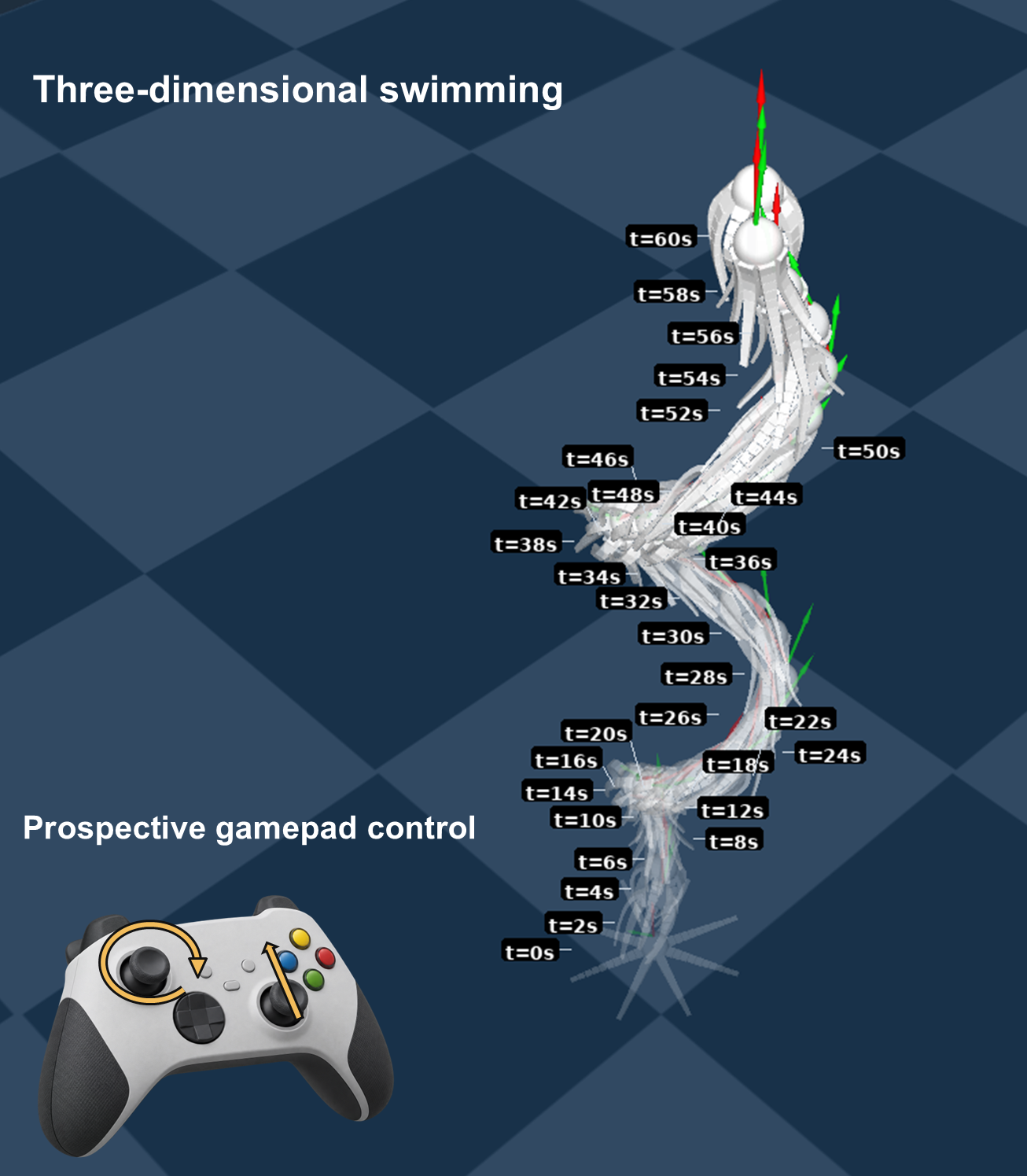}
    \caption{Three-dimensional swimming demonstration with Non-coordinated PPO with red and green arrows denote the commanded and measured world-frame velocities, respectively. Prospective interface in which circular joystick motion changes the lateral command direction while another input sets the upward component. The gamepad inset is a conceptual illustration generated using OpenAI's image-generation tool.}
    \label{fig:spiral_demo}
\end{figure}

\begin{samepage}
\section{Three-Dimensional Swimming Demonstration}
\label{sec:spiral_demo}

Non-coordinated PPO can also produce an ascending spiral under changing three-dimensional velocity commands Fig.~\ref{fig:spiral_demo}). We set the commanded velocity magnitude to $0.05\,\mathrm{m/s}$ and held each command for 2\,s. The afterimages show the resulting robot motion from 0 to 60\,s at 2-s intervals. Red and green arrows denote the commanded and measured world-frame velocities, respectively. This example illustrates how a sequence of velocity commands can generate a spatial complex swimming maneuver in simulation using this control architecture, and a concept that the robot could allow for controller-based control in the future.
\par
\end{samepage}

\begingroup
\widowpenalty=10000
A future gamepad interface could map circular motion to two joysticks. An operator could then supply velocity commands while the
policy controls arm actuation. The inset illustrates this proposed
interface. The present rollout uses scripted commands, motivating
subsequent evaluation of interactive control.
\par
\endgroup

\begin{samepage}
\section{CONCLUSIONS}
\label{sec:conclusion}

We presented a feedback swimming-control architecture
for an underwater robot with multiple underactuated soft
arms and showed that the tested eight-arm underactuated swimmer can control its swimming speed and direction through learned arm-specific amplitude modulation under a shared rhythmic drive, without prescribed inter-arm phase coupling.
\par
\end{samepage}

In simulation, Non-coordinated PPO achieved an 84.5\% success rate across 200 randomized three-dimensional velocity commands, based on the average velocity over the final 2~s, compared with 58.5\% for Coordinated PPO and 14.0\% for Direct PPO. Straight-line swimming and direction-change trials further supported command-dependent swimming under constant and changing velocity commands. These findings suggest that prescribed inter-arm phase coupling should be evaluated for the specific swimming task, rather than assumed necessary for effective multi-arm control, and direct learning approach may also not effective for above swimming task.

The findings are limited to the simulated eight-arm robot and its modeled fluid conditions, with randomized velocity tracking and representative maneuver trials providing the present evidence. Further evaluation on physical robots and other appendage arrangements and tracking metrics, is needed to be explored.






\section*{ACKNOWLEDGMENT}

OpenAI Codex assisted with drafting, restructuring, and language revision of text in the Abstract and Sections~\ref{sec:introduction}--\ref{sec:conclusion}, as well as literature searches and reference checking. OpenAI's image-generation tool was used to create the hydrodynamics illustration in the lower part of Fig.~\ref{fig:general}(b) and the gamepad illustration in Fig.~\ref{fig:spiral_demo}. These two images are conceptual illustrations rather than simulation outputs. The authors are responsible for the final content of this paper.



\IEEEtriggeratref{13}
\bibliographystyle{IEEEtran}
\bibliography{references}

@inproceedings{c1,
  author    = {Zhang, Bobing and Zhang, Yiyuan and Li, Yiming and Xuan, Sicheng and Ng, Hong Wei and Liufu, Yuliang and Tang, Zhiqiang and Laschi, Cecilia},
  title     = {Octopus-swimming-like robot with soft asymmetric arms},
  booktitle = {2025 IEEE 8th International Conference on Soft Robotics (RoboSoft)},
  pages     = {1--8},
  year      = {2025},
  doi       = {10.1109/robosoft63089.2025.11020919}
}

@article{c2,
  author  = {Wang, J. and Du, P. and Zhang, Y. and Xie, Z. and Laschi, C.},
  title   = {From shallow waters to Mariana Trench: A survey of bio-inspired underwater soft robots},
  journal = {Bioinspiration \& Biomimetics},
  volume  = {21},
  number  = {2},
  pages   = {021002},
  year    = {2026},
  doi     = {10.1088/1748-3190/ae3af2}
}

@article{c3,
  author  = {Li, Lei and Qin, Boyang and Gao, Wenzhuo and Li, Yanyu and Zhang, Yiyuan and Wang, Bo and Kong, Shihan and Wang, Jian and He, Dekui and Yu, Junzhi},
  title   = {Bioinspired underwater soft robots: From biology to robotics and back},
  journal = {npj Robotics},
  volume  = {4},
  number  = {1},
  pages   = {25},
  year    = {2026},
  doi     = {10.1038/s44182-026-00088-x}
}

@article{c4,
  author  = {Ijspeert, A. J.},
  title   = {Central pattern generators for locomotion control in animals and robots: A review},
  journal = {Neural Networks},
  volume  = {21},
  number  = {4},
  pages   = {642--653},
  year    = {2008},
  doi       = {10.1016/j.neunet.2008.03.014}
}

@misc{c10,
  author       = {Zhang, Chenrui and Zhang, Yiyuan and Ye, Yunfei and Chen, Junkai and Wang, Haozhe and Laschi, Cecilia},
  title        = {Unified structural-hydrodynamic modeling of underwater underactuated mechanisms and soft robots},
  howpublished = {arXiv preprint arXiv:2603.07939},
  year         = {2026},
  url       = {https://arxiv.org/abs/2603.07939},
  doi       = {10.48550/arXiv.2603.07939}
}

@article{c13,
  author  = {Sfakiotakis, M. and Kazakidi, A. and Tsakiris, D. P.},
  title   = {Octopus-inspired multi-arm robotic swimming},
  journal = {Bioinspiration \& Biomimetics},
  volume  = {10},
  number  = {3},
  pages   = {035005},
  year    = {2015},
  doi       = {10.1088/1748-3190/10/3/035005}
}

@article{c22,
  author={Van Dong Nguyen and Dinh Quoc Vo and Van Tu Duong and Huy Hung Nguyen and Tan Tien Nguyen},
  title={Reinforcement Learning-Based Optimization of Locomotion Controller Using Multiple Coupled {CPG} Oscillators for Elongated Undulating Fin Propulsion},
  journal={Mathematical Biosciences and Engineering},
  year={2022},
  volume={19},
  number={1},
  pages={738--758},
  doi={10.3934/mbe.2022033}
}

@article{c23,
  author={Imran Hameed and Xu Chao and David Navarro-Alarcon and Xingjian Jing},
  title={Deep Reinforcement Learning Enabling a {BCFbot} to Learn Various Undulatory Patterns},
  journal={Ocean Engineering},
  year={2025},
  volume={320},
  pages={120322},
  doi={10.1016/j.oceaneng.2025.120322}
}

@article{c25,
  author  = {Della Santina, Cosimo and Duriez, Christian and Rus, Daniela},
  title   = {Model-Based Control of Soft Robots: A Survey of the State of the Art and Open Challenges},
  journal = {IEEE Control Systems Magazine},
  volume  = {43},
  number  = {3},
  pages   = {30--65},
  year    = {2023},
  doi     = {10.1109/MCS.2023.3253419}
}

@article{c30,
  author    = {Armanini, Costanza and Farman, Madiha and Calisti, Marcello and Giorgio-Serchi, Francesco and Stefanini, Cesare and Renda, Federico},
  title     = {Flagellate Underwater Robotics at Macroscale: Design, Modeling, and Characterization},
  journal   = {IEEE Transactions on Robotics},
  volume    = {38},
  number    = {2},
  pages     = {731--747},
  year      = {2022},
  doi       = {10.1109/tro.2021.3094051}
}

@article{c31,
  author    = {AlShehhi, Aysha A. Samra and Armanini, Costanza and Mathew, Anup Teejo and Feliu-Talegon, Daniel and Renda, Federico},
  title     = {{Trident}: A Multi-Functional Soft Appendage for Underwater Locomotion and Grasping},
  journal   = {IEEE Robotics and Automation Letters},
  volume    = {10},
  number    = {1},
  pages     = {436--443},
  year      = {2025},
  doi       = {10.1109/lra.2024.3504238}
}

@inproceedings{c32,
  author    = {Kim, Sareum and Hughes, Josie},
  title     = {Octopus-Inspired Swimming Robot with Web-Coupled Multi-Arm Coordination},
  booktitle = {2025 IEEE 8th International Conference on Soft Robotics (RoboSoft)},
  pages     = {1--6},
  year      = {2025},
  doi       = {10.1109/robosoft63089.2025.11020935}
}

@article{c33,
  author    = {Wu, Mingxin and Afridi, Waqar Hussain and Wu, Jiaxi and Afridi, Rahdar Hussain and Wang, Kaiwei and Zheng, Xingwen and Wang, Chen and Xie, Guangming},
  title     = {Octopus-Inspired Underwater Soft Robotic Gripper with Crawling and Swimming Capabilities},
  journal   = {Research},
  volume    = {7},
  pages     = {0456},
  year      = {2024},
  doi       = {10.34133/research.0456}
}

@article{c34,
  author    = {Renda, Federico and Giorgio-Serchi, Francesco and Boyer, Frederic and Laschi, Cecilia and Dias, Jorge and Seneviratne, Lakmal},
  title     = {A unified multi-soft-body dynamic model for underwater soft robots},
  journal   = {The International Journal of Robotics Research},
  volume    = {37},
  number    = {6},
  pages     = {648--666},
  year      = {2018},
  doi       = {10.1177/0278364918769992}
}

@inproceedings{c35,
  author    = {Patterson, Zach J. and Sabelhaus, Andrew P. and Chin, Keene and Hellebrekers, Tess and Majidi, Carmel},
  title     = {An Untethered Brittle Star-Inspired Soft Robot for Closed-Loop Underwater Locomotion},
  booktitle = {2020 IEEE/RSJ International Conference on Intelligent Robots and Systems (IROS)},
  pages     = {8758--8764},
  year      = {2020},
  doi       = {10.1109/iros45743.2020.9341008}
}

\end{document}